\documentclass[conference]{IEEEtran}
\usepackage{cite}
\usepackage{amsmath,amssymb,amsfonts}
\usepackage{algorithmic}
\usepackage{graphicx}
\usepackage{textcomp}
\usepackage{balance}
\begin{document}

\title{Learning Burst-Aware Early Warning Models for Capacity Stress under AI Workload Surges in Hyperscale Data Centers}

\author{%
\begin{tabular}{ccc}
\begin{tabular}{c}
\textbf{Zihan Yu\textsuperscript{*}} \\
\textit{College of Professional Studies} \\
\textit{Northeastern University} \\
Boston, USA \\
yu.zihan1@northeastern.edu
\end{tabular}
&
\begin{tabular}{c}
\textbf{Xianling Zeng} \\
\textit{College of Engineering} \\
\textit{Northeastern University} \\
Boston, USA \\
zeng.xian@northeastern.edu
\end{tabular}
&
\begin{tabular}{c}
\textbf{Zhiming Xue} \\
\textit{College of Engineering} \\
\textit{Northeastern University} \\
Boston, USA \\
xue.zh@northeastern.edu
\end{tabular}
\\[-0.2ex]
\multicolumn{3}{c}{%
\begin{tabular}{cc}
\begin{tabular}{c}
\textbf{Yalun Qi} \\
\textit{Khoury College of Computer Sciences} \\
\textit{Northeastern University} \\
Boston, USA \\
qi.yal@northeastern.edu
\end{tabular}
&\hspace{0.6cm}
\begin{tabular}{c}
\textbf{Sichen Zhao} \\
\textit{College of Engineering} \\
\textit{Northeastern University} \\
Boston, USA \\
zhao.siche@northeastern.edu
\end{tabular}
\end{tabular}}
\end{tabular}
}

\maketitle

\begin{abstract}
The rapid growth of large-scale AI workloads, particularly Large Language Model (LLM) training and inference, is fundamentally reshaping the operational dynamics of hyperscale data centers. Unlike traditional cloud workloads, AI-driven jobs exhibit bursty, high-intensity, and rapidly shifting resource demands, often leading to sudden capacity stress that cannot be effectively handled by reactive threshold-based mechanisms. In this paper, we propose a deployment-oriented, burst-aware early warning framework for proactive capacity stress prediction under AI workload surges. We formulate the problem as a high-recall forecasting task over multivariate telemetry windows, with the explicit goal of enabling operational intervention before system degradation occurs. The proposed framework integrates workload intensity, temporal variation, and system pressure signals, and employs a lightweight tree-based learning model to capture nonlinear interactions in highly imbalanced environments. To evaluate the system under realistic conditions, we introduce an AI workload surge injection methodology that simulates burst-driven demand patterns observed in large-scale AI systems. Experimental results show that AI workload surges increase the capacity stress rate by 37.9\%, highlighting the urgency of proactive mitigation. Our XGBoost-based model achieves an ROC AUC of 0.697 and an AP of 0.670, significantly outperforming baseline methods. Under deployment-oriented threshold selection, the framework achieves a Recall of 0.914, enabling the detection of the majority of stress-prone periods with acceptable false-alarm cost. Beyond predictive performance, we show how the proposed framework can be integrated into operational control loops to support proactive actions such as workload throttling and resource scaling. Our results highlight the practical value of high-recall, learning-based early warning systems in enabling resilient and adaptive data center operations in the era of AI-driven workloads.
\end{abstract}


\section{Introduction}
The rapid proliferation of large-scale AI workloads, particularly Large Language Model (LLM) training and inference, is fundamentally changing how hyperscale data centers operate\cite{ref2}. Unlike traditional cloud applications with relatively stable demand patterns, AI workloads introduce bursty, high-intensity resource consumption that can rapidly push systems into capacity stress states within short time horizons. These burst-driven dynamics create significant challenges for maintaining system stability, as resource contention can escalate quickly and propagate across shared infrastructure.

Existing capacity management approaches are predominantly reactive, relying on static thresholds or post-hoc scaling mechanisms that respond only after resource pressure has already materialized. Under AI workload surges, such reactive strategies are often too slow to prevent performance degradation and Service Level Objective (SLO) violations. Moreover, traditional workload prediction methods are not designed to capture rare but high-impact stress events, particularly under highly dynamic and imbalanced conditions.

This mismatch highlights a critical need for proactive, deployment-oriented early warning systems that can identify stress-prone periods before they fully manifest, enabling timely operational intervention. Unlike conventional prediction tasks that prioritize overall accuracy, early warning systems in data center operations must emphasize high recall, as missing a stress event can lead to significant service disruption, while false positives typically incur relatively low operational cost.

To address this challenge, we propose a deployment-oriented, burst-aware early warning framework for proactive capacity stress prediction under AI workload surges. We formulate the problem as a high-recall forecasting task over multivariate telemetry windows, where the goal is to detect future stress-prone periods with sufficient lead time to enable intervention. The proposed framework integrates workload intensity, temporal variation, and system pressure signals to capture the characteristic dynamics of burst-driven workloads. To support realistic evaluation, we introduce an AI workload surge injection methodology that simulates high-intensity demand patterns observed in large-scale AI systems. We employ a lightweight tree-based learning model to capture nonlinear relationships in heterogeneous telemetry data, enabling efficient and practical deployment.

Experimental results demonstrate that AI workload surges significantly increase the frequency and volatility of capacity stress events, underscoring the importance of proactive mitigation. Under a deployment-oriented threshold selection, the proposed framework achieves high recall in identifying stress-prone periods, providing actionable early warning signals that can support operational decision-making. Beyond predictive performance, the framework can be integrated into system control loops to enable proactive actions such as workload throttling, resource scaling, and load redistribution.

The main contributions of this paper are as follows:
\begin{itemize}
\item We identify AI workload surges as a key driver of burst-driven capacity stress in hyperscale data centers, and highlight the fundamental mismatch between AI-era workload dynamics and traditional reactive capacity management.
\item We propose a deployment-oriented, burst-aware early warning framework that integrates workload intensity, temporal variation, and system pressure signals to proactively identify stress-prone periods under highly dynamic and imbalanced conditions.
\item We introduce an AI workload surge injection methodology to emulate realistic burst scenarios, enabling systematic evaluation of early warning systems under conditions that are not observable in publicly available datasets.
\item We demonstrate that a high-recall, threshold-aware decision framework can effectively support proactive operational interventions, providing practical value beyond traditional accuracy-oriented prediction models.
\end{itemize}

\section{Related Work}
\subsection{Data Center Capacity Management}
Traditional capacity management in hyperscale environments has focused on long-term hardware provisioning and steady-state resource allocation \cite{ref1}. Recent work has explored dynamic resource management to improve utilization. For instance, FIRM proposes a fine-grained resource management framework for SLO-oriented microservices \cite{ref5}, and Meta's ReBalancer addresses global resource allocation across millions of servers \cite{ref6}. Capacity variation penalties have also been studied in cloud resource management \cite{ref3}. However, these approaches often assume relatively stable workload patterns and struggle with the extreme burstiness of modern AI workloads.

\subsection{Anomaly Detection and Workload Prediction}
Machine learning has been widely applied to anomaly detection and workload prediction in cloud environments. Several studies have utilized ARIMA, LSTM, and gradient boosting methods for predicting server load and energy consumption \cite{ref7,ref8}. While these models perform well for general cloud workloads, predicting capacity stress---a rare and extreme event---presents unique challenges related to severe class imbalance and rapid temporal dynamics. Furthermore, regime-dependent analysis has demonstrated that time-series relationships can shift markedly across distinct operational states---a finding directly applicable to the Normal vs. AI-Surge regimes studied in this work \cite{ref10}.

\subsection{Handling Microbursts}
Network microbursts in data centers have been extensively studied. Systems like Trumpet \cite{ref9} provide high-resolution monitoring to detect short-lived network anomalies. Our work extends this concept from the network layer to holistic, system-wide capacity stress, focusing on the macroscopic impact of AI workload surges across compute and memory resources. Unlike prior work that focuses on steady-state anomaly detection or network-level microbursts, our work explicitly targets deployment-oriented early warning under AI workload surges, prioritizing high-recall decision making under severe class imbalance.

\section{Methodology}
\subsection{Problem Formulation}
We formulate the task as a deployment-oriented early warning problem over multivariate telemetry windows. Rather than optimizing generic classification accuracy, the objective is to identify future stress-prone periods early enough to enable operational intervention before severe degradation occurs. This setting emphasizes early detection of rare but high-impact events, which is critical for maintaining system stability under burst-driven AI workload surges. Let $\mathbf{X}_t$ represent the multivariate telemetry vector at time window $t$, as defined in (
\ref{eq:x}):
\begin{equation}
\mathbf{X}_t = [x_t^{(1)}, x_t^{(2)}, \ldots, x_t^{(d)}]^\top \in \mathbb{R}^d
\label{eq:x}
\end{equation}

Each feature vector is constructed over a fixed-length window and concatenates multiple resource metrics into a unified representation. The input feature vector consists of multivariate telemetry signals, including CPU utilization, memory usage, and network throughput, which collectively characterize the system's resource dynamics. These features are designed to capture both sustained load patterns and transient spikes that may lead to capacity stress. All features are normalized to ensure stable model training and to mitigate scale differences across heterogeneous metrics. Temporal dependencies are captured through lag-based and rolling-window features, which enable the model to detect short-horizon variations and burst patterns in system behavior. This burst-aware representation allows the framework to respond not only to sustained high utilization, but also to rapid changes that often precede capacity stress under AI workload surges.

Let $Y_{t+k}$ be the binary stress label at a future time window $t+k$, as shown in (\ref{eq:y}):
\begin{equation}
Y_{t+k} \in \{0,1\}, \qquad k \geq 1
\label{eq:y}
\end{equation}

Our goal is to learn a mapping function $f$ that estimates the probability of impending capacity stress, as formalized in (\ref{eq:f}):
\begin{equation}
f: \mathbf{X}_t \longrightarrow \hat{p}_t = P(Y_{t+k}=1\mid \mathbf{X}_t)
\label{eq:f}
\end{equation}

At inference time, the early warning decision is made by comparing the predicted probability against a classification threshold $\tau$, as shown in (\ref{eq:decision}):
\begin{equation}
\hat{Y}_{t+k} = \mathbf{1}[\hat{p}_t \geq \tau], \qquad \tau = 0.1
\label{eq:decision}
\end{equation}

In practice, the prediction threshold is not selected solely based on threshold-independent metrics such as ROC AUC. Instead, it is determined according to operational requirements, where missing a stress event (false negative) is significantly more costly than issuing a false alarm. Therefore, the model is designed to operate under a high-recall regime, ensuring that the majority of stress-prone periods can be identified in advance. The choice of the threshold $\tau$ governs the trade-off between precision and recall. In the context of early warning, the threshold is selected to satisfy a high-recall objective, prioritizing the detection of rare but high-impact stress events over minimizing false positives. The specific threshold value is determined empirically based on validation performance and is discussed in detail in Section~\ref{sec:threshold}.

\subsection{Data Collection and AI Surge Injection}
To emulate burst-driven infrastructure conditions, we construct a window-level telemetry dataset and introduce an AI workload surge injection procedure. The data is aggregated into 5-minute windows over approximately 30 days. We distinguish a Normal Period representing baseline cloud operations from an AI Surge Period where high-intensity, bursty demand patterns are injected to emulate large-scale AI training and inference behavior.

We acknowledge that this injection model simplifies real LLM workloads. It omits multi-phase effects such as epoch-wise batch variation, collective communication, checkpoint I/O bursts, and inference request volatility, which may underestimate the temporal complexity of real AI-driven stress events. Future work should calibrate these injection parameters using traces from production-scale AI clusters.

\subsection{Model Architecture: XGBoost}
In this work, we adopt a lightweight and deployment-friendly tree-based model as the predictive engine of the proposed framework. This design choice prioritizes inference efficiency, robustness under class imbalance, and ease of integration into real-time monitoring systems. Tree-based models are particularly well suited to heterogeneous infrastructure telemetry, as they can capture nonlinear interactions among workload intensity, temporal variation, and system pressure signals without requiring strong distributional assumptions. This makes them effective for modeling burst-driven dynamics under AI workload surges. We implement the predictive engine using XGBoost \cite{ref4}, a gradient-boosted tree model that builds an ensemble of decision trees in a stage-wise manner. This allows the model to capture complex nonlinear relationships while maintaining low inference latency, which is critical for deployment in operational environments. The regularized objective function is given in (\ref{eq:xgb}):
\begin{equation}
\begin{aligned}
\mathcal{L} &= \sum_{i=1}^{n} \ell(y_i, \hat{y}_i^{(T)}) + \sum_{t=1}^{T} \Omega(f_t), \\
\Omega(f) &= \gamma\lvert\mathrm{leaves}\rvert + \frac{1}{2}\lambda\lVert w \rVert^2
\end{aligned}
\label{eq:xgb}
\end{equation}
This consideration is consistent with recent hardware-aware model co-design work \cite{ref11}, which demonstrates that matching model topology to inference substrate constraints is critical for achieving real-time performance in production environments.

XGBoost is particularly suitable for this early warning setting for several reasons. First, it effectively handles heterogeneous and partially correlated system metrics without requiring extensive feature normalization. Second, it supports imbalance-aware training through built-in weighting mechanisms, which is critical for detecting rare stress events. Third, its inference efficiency enables low-latency prediction, making it practical for deployment in real-time monitoring systems. Finally, the model provides feature importance scores that help interpret key drivers of predicted capacity stress, offering additional insight for system operators. To further contextualize its performance, we compare XGBoost with baseline models including Logistic Regression and LightGBM in the experimental evaluation.

Modern deep learning-based time-series models such as TFT and N-BEATS are not the focus here because they require more labeled stress windows and larger inference budgets than available in our deployment setting. Their training warm-up, latency, and memory costs are less compatible with real-time monitoring on structured telemetry windows, for which gradient-boosted trees are a better fit. We leave these models to future work with larger datasets and stronger inference infrastructure.

\section{Experimental Evaluation}
\subsection{Dataset}
We conduct experiments using a real-world workload trace derived from the Google Cluster Workload Traces (2019), which represents production-scale resource usage in large compute clusters. The trace contains fine-grained job-level telemetry, including CPU, memory, and scheduling metadata, making it suitable for studying capacity stress dynamics in data center environments.

In this work, we extract temporal workload signals and construct window-level features to characterize system load variations and capacity stress patterns. Compared to synthetic benchmarks, the use of a real-world production trace improves the practical relevance and generalizability of the framework.

At the same time, we acknowledge that the Google Cluster Workload Traces predate the widespread deployment of modern large-scale LLM training and inference workloads, so the base trace does not natively capture the full resource signature of contemporary AI jobs. To partially compensate for this gap, we synthetically inject AI workload surge patterns calibrated to observed properties of large-scale AI clusters. Accordingly, the present evaluation should be interpreted as a proof-of-concept on a well-established benchmark, with future validation on newer AI-centric production traces remaining an important next step.

\begin{table}[htbp]
\caption{Key Features Used in the Model}
\label{tab:features}
\centering
\begin{tabular}{|p{0.18\columnwidth}|p{0.28\columnwidth}|p{0.42\columnwidth}|}
\hline
\textbf{Category} & \textbf{Features} & \textbf{Description} \\
\hline
Resource Usage & {CPU utilization, memory usage trends} & Reflects system load and capacity utilization \\
\hline
Temporal Signals & Workload variance, moving average & Captures dynamic changes over time \\
\hline
System Behavior & Job arrival rate, scheduling density & Represents workload intensity and scheduling patterns \\
\hline
Derived Metrics & Capacity stress indicator, change rate & Indicates potential system stress conditions \\
\hline
\end{tabular}
\end{table}

\subsection{Experimental Setup}
All experiments are conducted using the preprocessed Google cluster workload dataset. The data is segmented into fixed temporal windows to capture workload dynamics over time. Continuous features are normalized to ensure stable model training and to mitigate scale differences across heterogeneous metrics. We adopt XGBoost as the primary prediction model due to its robustness and strong performance on structured data. The dataset is split into training and testing sets using a chronological split to avoid temporal leakage. Model performance is evaluated using standard classification metrics including precision, recall, and F1-score. Hyperparameters are tuned using a validation set to balance recall and precision under class imbalance conditions.

\subsection{Impact of AI Workload Surges}
\begin{figure}[htbp]
\centering
\includegraphics[width=0.9\columnwidth]{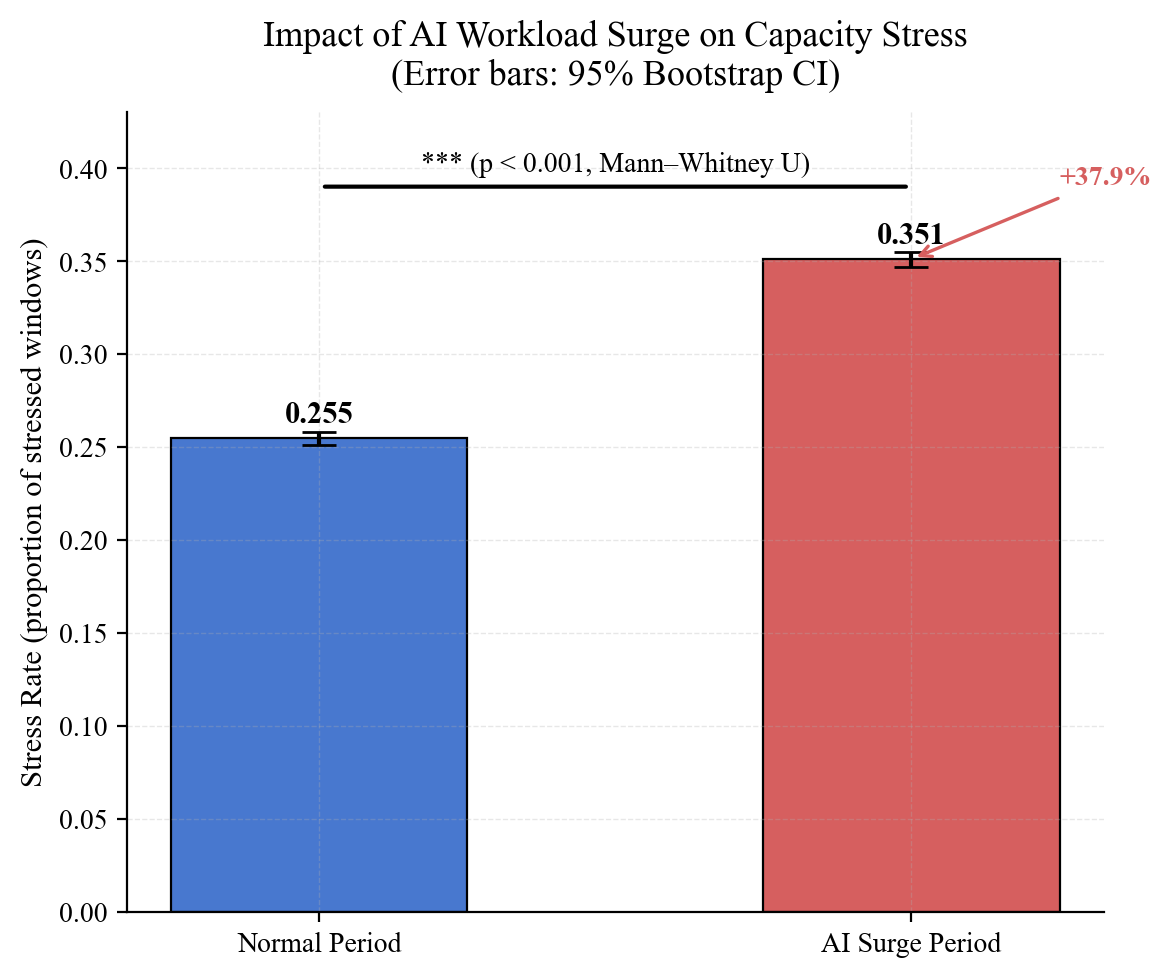}
\caption{Comparison of capacity stress rates between the Normal Period and the AI Surge Period. Error bars represent 95\% bootstrap confidence intervals. Statistical significance is confirmed by a Mann-Whitney U test ($***\ p < 0.001$).}
\label{fig:stressrate}
\end{figure}

We first quantify the impact of AI workloads on data center capacity. Figure~\ref{fig:stressrate} illustrates the proportion of time windows experiencing capacity stress during the Normal Period versus the AI Surge Period.

During the Normal Period, the baseline stress rate is 0.255. Following the injection of AI workloads, the stress rate jumps significantly to 0.351, representing a relative increase of 37.9\%. This finding underscores the severe strain that AI workloads place on existing infrastructure and highlights the necessity of proactive management.

\begin{figure}[htbp]
\centering
\includegraphics[width=\columnwidth]{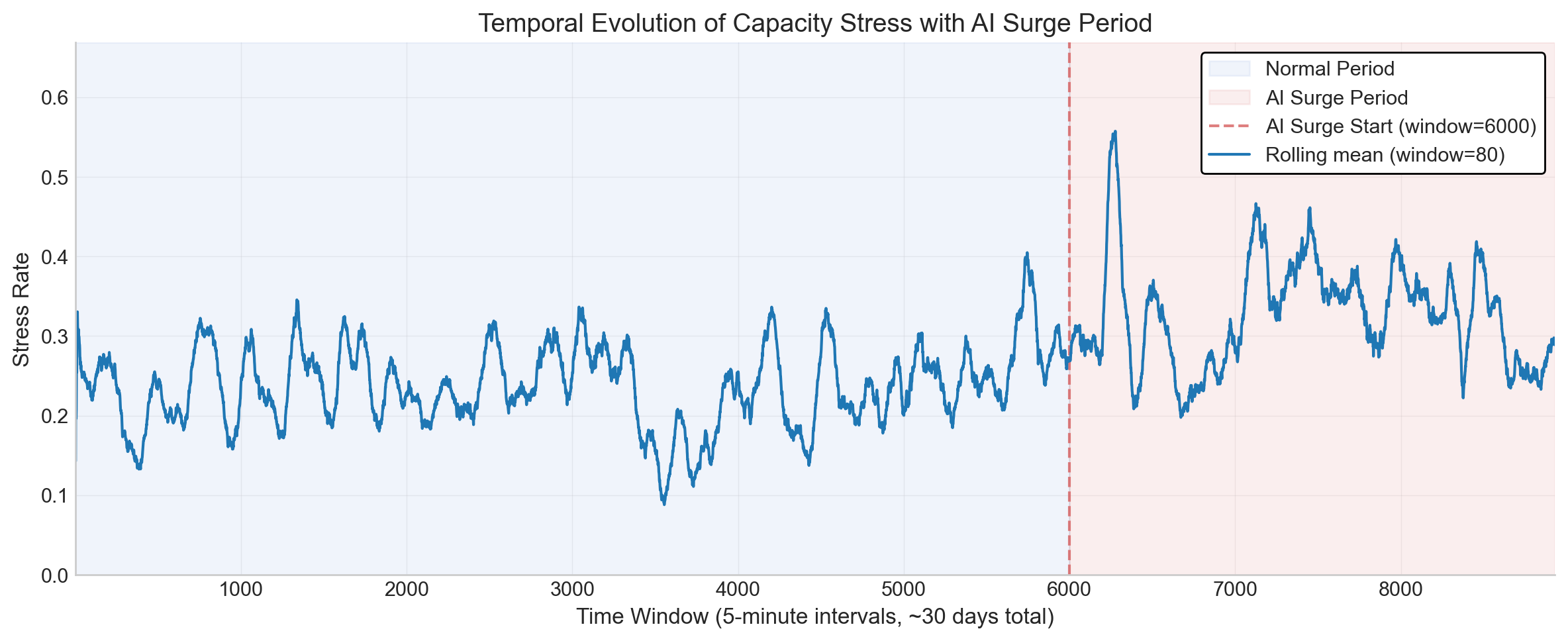}
\caption{Rolling mean of capacity stress rate over $\sim$30 days (5-minute windows). The blue region denotes the Normal Period; the red region denotes the AI Surge Period. The red dashed line marks the AI Surge onset at window 6,000.}
\label{fig:rolling}
\end{figure}

Figure~\ref{fig:rolling} visualizes the temporal evolution of capacity stress over the full observation period. The red dashed line marks the onset of the AI Surge Period (at time window 6,000). The rolling mean clearly demonstrates a sustained elevation and increased volatility in stress levels following the surge.

These observations highlight the importance of modeling burst-driven workload dynamics for accurate stress prediction.

\subsection{Model Performance}
We evaluate the proposed framework using three representative models: Logistic Regression, LightGBM, and XGBoost. Model thresholds are selected to satisfy a minimum recall requirement of 0.90, reflecting the priority of minimizing missed stress events.

\begin{table}[htbp]
\caption{Model Performance Comparison for Early Warning of Capacity Stress Under AI Workload Surges}
\label{tab:perf}
\centering
\resizebox{\columnwidth}{!}{%
\begin{tabular}{|l|c|c|c|c|c|}
\hline
\textbf{Model} & \textbf{ROC-AUC} & \textbf{AP} & \textbf{Precision} & \textbf{Recall} & \textbf{F1} \\
\hline
Logistic Regression & 0.576 & 0.129 & 0.083 & 0.977 & 0.153 \\
\hline
LightGBM & 0.532 & 0.097 & 0.085 & 0.909 & 0.156 \\
\hline
XGBoost & 0.697 & 0.670 & 0.525 & 0.914 & 0.667 \\
\hline
\end{tabular}%
}
\end{table}

As shown in Table~\ref{tab:perf}, Logistic Regression achieves high recall but suffers from low precision, while LightGBM underperforms on ROC AUC despite its greater model complexity. On this highly imbalanced dataset, LightGBM appears less stable than XGBoost for ranking sparse positive windows, and its boosting behavior is less sensitive to weak positive-class signals.

We further note that the baseline comparison is intentionally scoped to lightweight models that share XGBoost's deployment profile. Comparisons with heavier deep learning baselines are deferred to future work with larger labeled datasets and dedicated inference infrastructure.

XGBoost therefore achieves the most favorable balance between predictive performance and deployment practicality under severe class imbalance.

The results in Table~\ref{tab:perf} are reported under a fixed high-recall operating constraint (Recall $\geq 0.90$), while the ROC and PR curves evaluate model performance across all possible thresholds.

\begin{figure}[htbp]
\centering
\includegraphics[width=0.9\columnwidth]{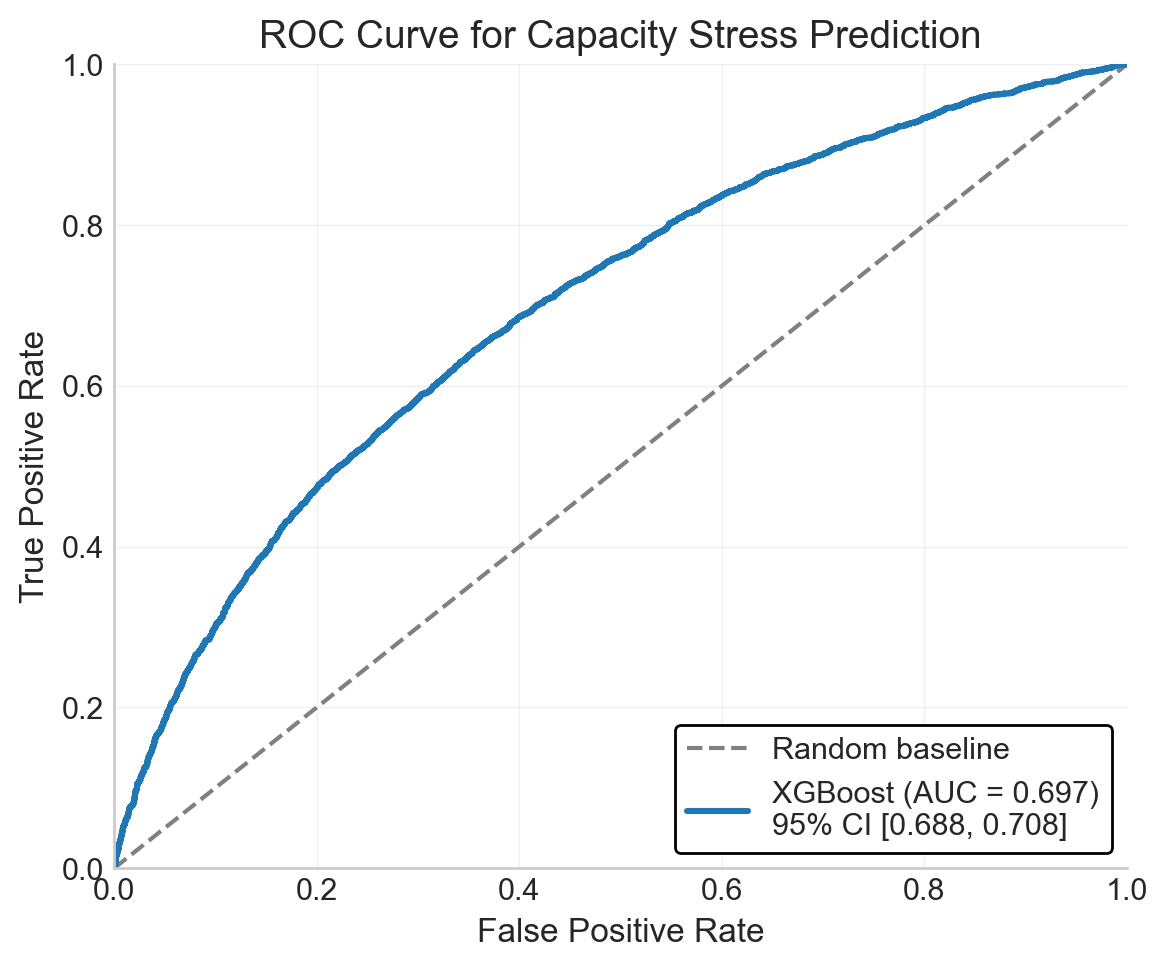}
\caption{ROC curve for the XGBoost model. The model achieves an AUC of 0.697 (95\% bootstrap CI [0.688, 0.708]), substantially outperforming the random baseline (AUC = 0.5).}
\label{fig:roc}
\end{figure}

\begin{figure}[htbp]
\centering
\includegraphics[width=\columnwidth]{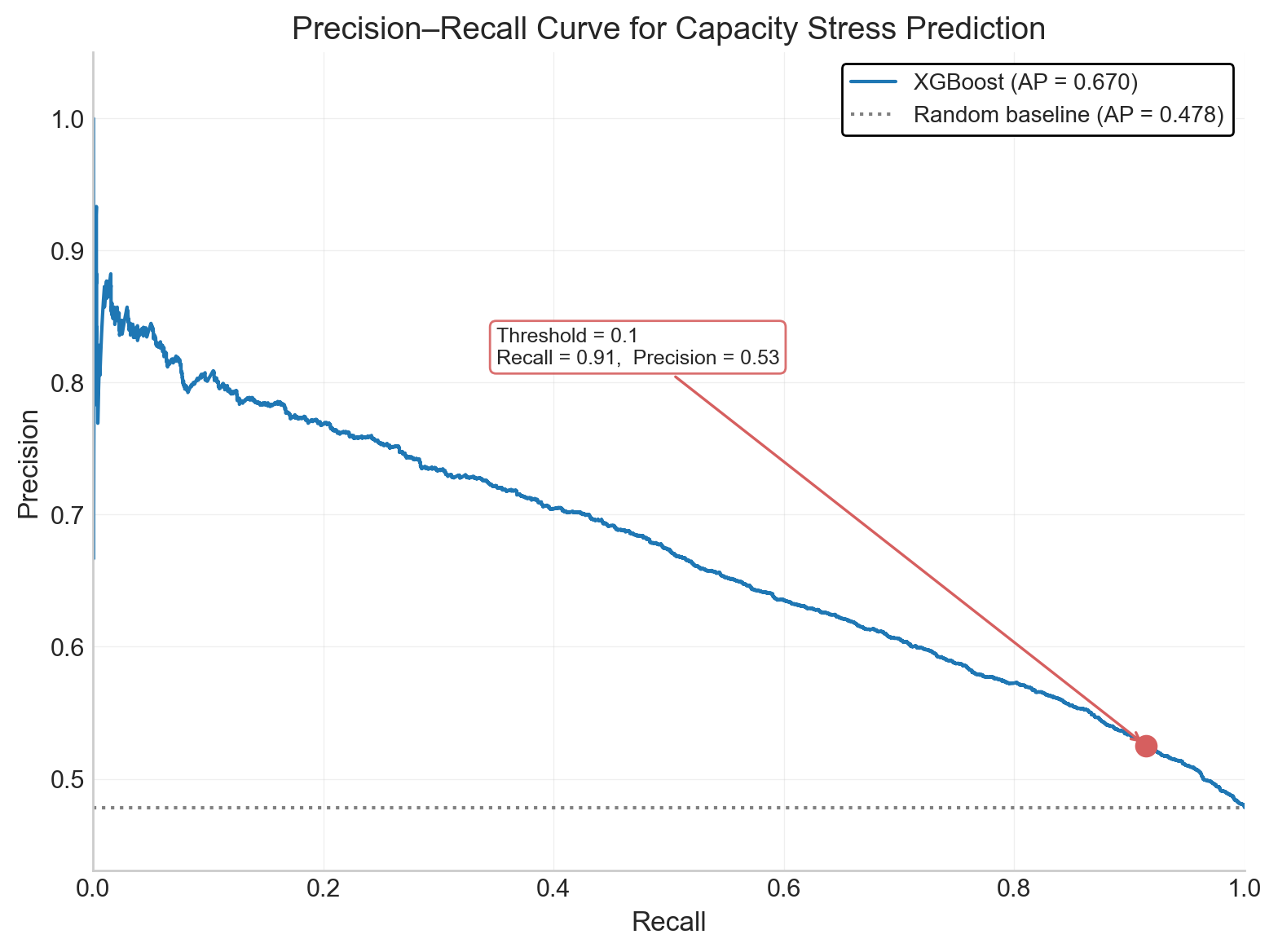}
\caption{{Precision-recall curve for the XGBoost model (AP = 0.670) versus the random baseline (AP = 0.478). The red dot marks the operating point at the selected threshold $\tau = 0.1$, where Precision = 0.525 and Recall = 0.914. The jagged region near Recall = 0 reflects low-threshold instability caused by sparse positive predictions.}}
\label{fig:pr}
\end{figure}

As shown in Figure~\ref{fig:roc}, the XGBoost model achieves an ROC AUC of 0.697. However, ROC curves can be overly optimistic on imbalanced datasets. We therefore also examine the PR curve (Figure~\ref{fig:pr}).

The model achieves an Average Precision (AP) of 0.670, significantly outperforming the random baseline (AP = 0.478). The PR curve highlights the inherent trade-off in early warning systems: achieving high recall inevitably comes at the cost of lower precision.

The instability observed near Recall = 0 is attributable to sparse positive predictions at very low thresholds and does not affect the operating region of interest.

\subsection{{Threshold Selection for Early-Warning Deployment}}
\label{sec:threshold}
In data center operations, the cost of a false negative (failing to predict a stress event, leading to an outage) is typically much higher than the cost of a false positive (unnecessarily throttling background tasks). Therefore, an effective early warning system must prioritize recall. This mirrors a broader principle in intelligent system design: knowing when to trigger an action is insufficient without a well-defined strategy for how to act upon it \cite{ref14}.

\begin{figure}[htbp]
\centering
\includegraphics[width=\columnwidth]{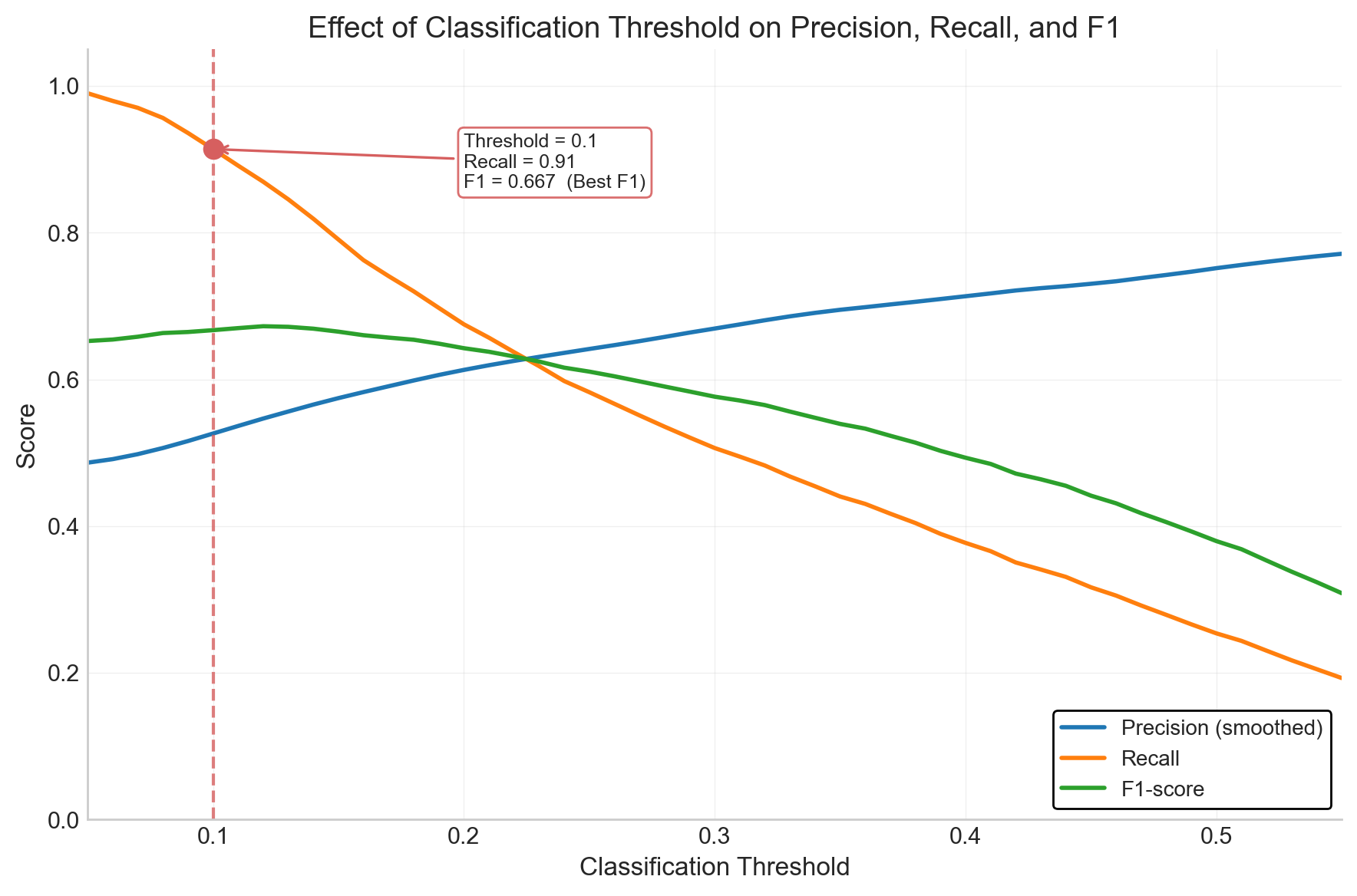}
\caption{{Effect of the classification threshold on precision (smoothed), recall, and F1-score. The red dashed line marks the selected threshold $\tau = 0.1$, which achieves a recall of 0.914 and an F1-score of 0.667 at the selected operating threshold.}}
\label{fig:threshold}
\end{figure}

Figure~\ref{fig:threshold} illustrates the effect of varying the classification threshold on all three metrics.
At the selected threshold of $\tau = 0.1$, the model achieves a recall of 0.914, meaning the system successfully anticipates 91.4\% of all capacity stress events. The corresponding precision of 0.525 implies a moderate false-alarm rate, which is an acceptable trade-off given the low cost of preventive actions (e.g., pausing batch jobs) relative to the cost of an undetected outage.

\subsection{{Interpretability Analysis}}
\label{sec:shap}
We use TreeSHAP to explain XGBoost predictions at both global and instance levels. Figure 6 shows that memory-demand and failure-rate features dominate, while Figure 7 shows monotonic directionality: high memory demand and elevated failure rates push predictions toward stress, whereas low values reduce stress probability. These patterns indicate that the model learned physically meaningful relationships and help operators distinguish memory-constrained alerts from compute-bound bursts.

\begin{figure}[t]
\centering
\includegraphics[width=0.46\columnwidth]{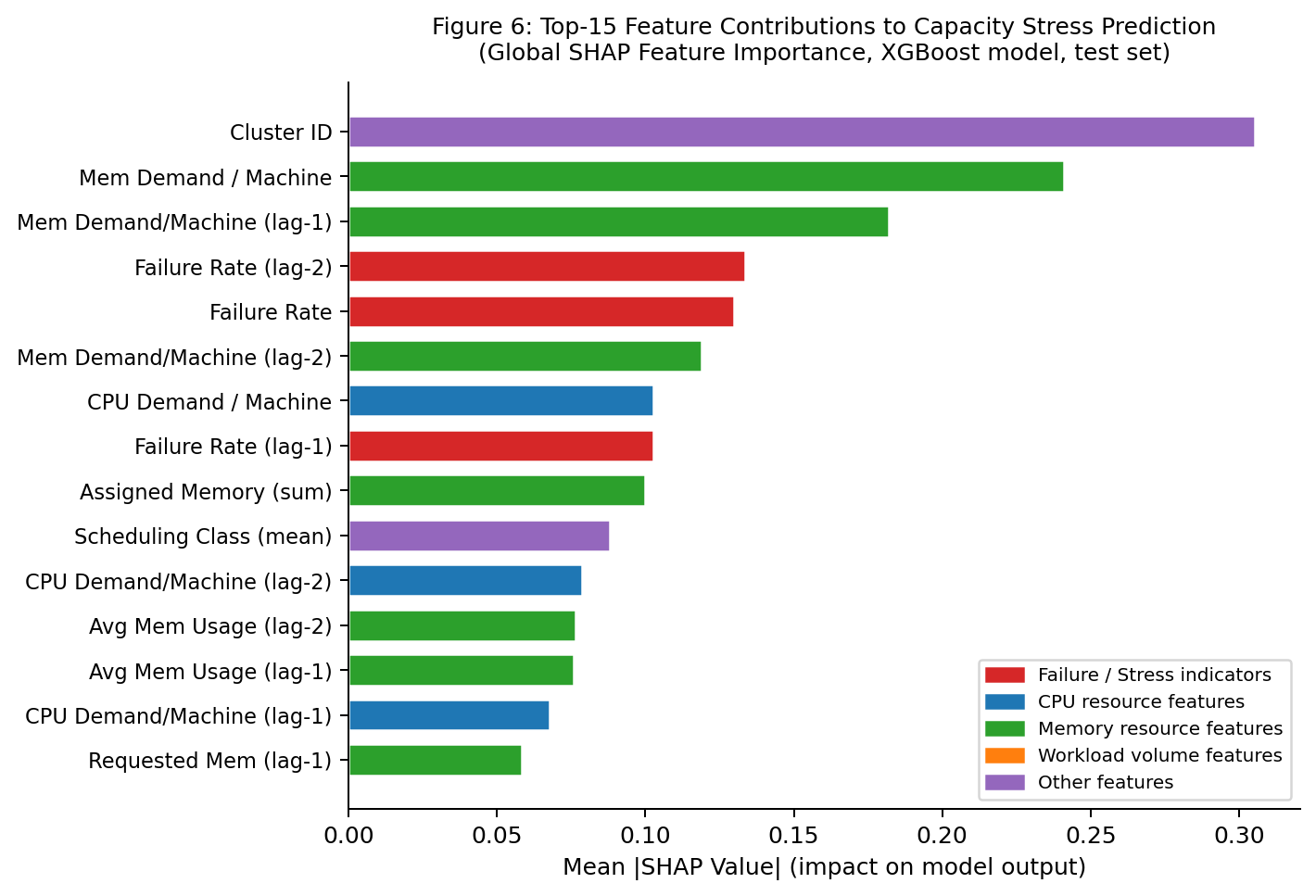}
\caption{Global SHAP feature importance for the XGBoost early warning model. Memory-demand and failure-rate features dominate the ranking.}
\label{fig:shap-global}
\end{figure}

\begin{figure}[t]
\centering
\includegraphics[width=0.46\columnwidth]{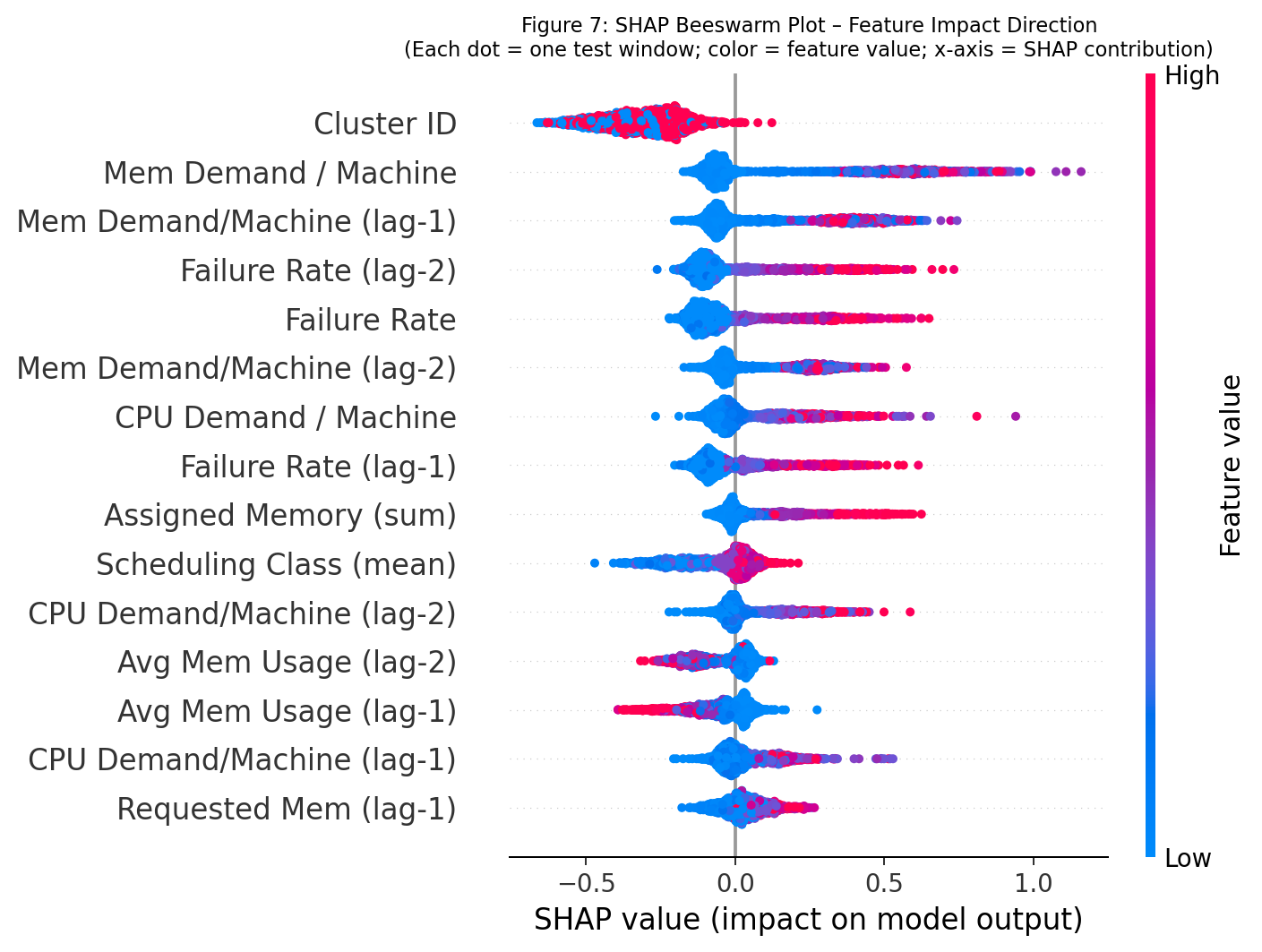}
\caption{SHAP beeswarm plot for individual predictions. High memory demand and elevated failure rates consistently push predictions toward stress, while low values reduce stress probability.}
\label{fig:shap-beeswarm}
\end{figure}

\noindent Note on Cluster ID. Its high SHAP importance likely reflects cluster-level fixed effects rather than a causal stress mechanism. Excluding Cluster ID, memory-demand and failure-rate features remain the main precursors identified in Section~III.

\section{Discussion}
These results highlight a fundamental shift from reactive resource management to predictive, learning-driven control in AI-centric data center operations. By operating at a threshold of $\tau = 0.1$, our system provides data center operators with a critical lead time to take preventive action. From a system perspective, the proposed early warning mechanism can be integrated into automated orchestration frameworks to enable proactive mitigation. When a warning is triggered, automated orchestration systems can:
\begin{enumerate}
\item Pause or migrate low-priority batch jobs (e.g., data backups, log processing) to free up resources immediately.
\item Proactively spin up additional virtual machines or allocate reserve hardware before the AI burst fully materializes.
\item Implement dynamic rate limiting on incoming API requests to prevent cascading failures.
\end{enumerate}

While the precision of 0.525 implies a moderate false-alarm rate, the mitigation actions described above are generally low-cost and non-disruptive to user-facing services. Thus, the trade-off heavily favors the high recall achieved by our model. This highlights the importance of integrating predictive models into operational control loops, shifting from passive monitoring to active system optimization.

\textit{Limitations.} The current evaluation relies on synthetically injected AI surge patterns. While the injection methodology is grounded in observed characteristics of real AI workloads, validation on production traces from live hyperscale environments remains an important direction for future work. Furthermore, the moderate AUC of 0.697 suggests that richer feature engineering---incorporating thermal metrics, power draw, and inter-rack network telemetry---could further improve predictive accuracy. More broadly, our model faces a generalization challenge analogous to the memorization-versus-reasoning tension documented in recent LLM studies \cite{ref13}: the learned feature correlations may reflect surface-level co-occurrences in the training distribution rather than robust causal signals, motivating future work on distribution-shift robustness and continual learning for capacity stress prediction.

\section{Conclusion}
The rapid growth of AI workloads is fundamentally redefining the operational paradigm of hyperscale data centers. In this paper, we quantified the impact of AI workload surges, demonstrating a statistically significant 37.9\% increase in capacity stress rates. To address this challenge, we proposed a deployment-oriented, burst-aware early warning framework for proactive capacity stress prediction. By formulating the problem as a high-recall forecasting task and selecting an operationally meaningful decision threshold, the proposed framework successfully identifies the majority of stress-prone periods in advance (Recall = 0.914), enabling timely and practical intervention. Unlike conventional accuracy-oriented prediction approaches, our method emphasizes decision effectiveness under class imbalance, aligning with real-world operational priorities.

Future work will focus on validating the framework on production-scale telemetry, extending the feature space with additional system signals such as power and thermal metrics, and integrating predictive models into orchestration frameworks (e.g., Kubernetes) for closed-loop mitigation. To further scale deployment across millions of servers, a cascade architecture inspired by the Filter-And-Refine paradigm \cite{ref12} could reduce inference overhead while preserving high recall. More broadly, the challenge of robust generalization remains related to the memorization-versus-reasoning tension highlighted in recent LLM work \cite{ref13}, while pattern-aware tool-integrated reasoning agents \cite{ref14} may improve adaptive response selection in future orchestration pipelines.

\enlargethispage{4\baselineskip}

\end{document}